\documentclass[letterpaper]{article} % DO NOT CHANGE THIS
\usepackage{aaai25}  % DO NOT CHANGE THIS
\usepackage{times}  % DO NOT CHANGE THIS
\usepackage{helvet}  % DO NOT CHANGE THIS
\usepackage{courier}  % DO NOT CHANGE THIS
\usepackage[hyphens]{url}  % DO NOT CHANGE THIS
\usepackage{graphicx} % DO NOT CHANGE THIS
\usepackage{natbib}  % DO NOT CHANGE THIS AND DO NOT ADD ANY OPTIONS TO IT
\usepackage{caption} % DO NOT CHANGE THIS AND DO NOT ADD ANY OPTIONS TO IT
\usepackage{algorithm}
\usepackage{algorithmic}

\usepackage{xspace}
\newcommand{\mgtta}{MGTTA\xspace}
\usepackage{newfloat}
\usepackage{listings}
\DeclareCaptionStyle{ruled}{labelfont=normalfont,labelsep=colon,strut=off} % DO NOT CHANGE THIS
\floatstyle{ruled}
\newfloat{listing}{tb}{lst}{}
\floatname{listing}{Listing}
\usepackage{multicol}
\usepackage{multirow}
\usepackage{makecell}  % 导入 makecell 宏包

\let\emph\textit % 解决斜体命令\emph失效的bug
\usepackage{xspace}
\newcommand{\ie}{{\emph{i.e.}}\xspace}

\newcommand{\eg}{{\emph{e.g.}}\xspace}
\newcommand{\wrt}{{\emph{w.r.t.}}\xspace}
\newcommand{\etc}{etc.}

\usepackage{booktabs} 
\usepackage{threeparttable}
\usepackage{amssymb}

\usepackage{xspace}
\newcommand{\methodname}{MGTTA\xspace}
\usepackage{color}

\usepackage{amsmath} % 公式

\usepackage{xcolor}

\def\rh{\textcolor{black}}
\def\dq{\textcolor{black}}

\title{Learning to Generate Gradients for Test-Time Adaptation \\ via Test-Time Training Layers}
\author{
    Qi Deng\thanks{Equal Contribution}\textsuperscript{\rm 1},
    Shuaicheng Niu$^\ast$\textsuperscript{\rm 2},
    Ronghao Zhang\textsuperscript{\rm 1},
    Yaofo Chen\textsuperscript{\rm 1},\\
    Runhao Zeng\textsuperscript{\rm 3}\thanks{Corresponding Author},
    Jian Chen\textsuperscript{\rm 1}$^\dag$,
    Xiping Hu\textsuperscript{\rm 3}
}
\affiliations{
    \textsuperscript{\rm 1}South China University of Technology,
    \textsuperscript{\rm 2}Nanyang Technological University,\\
    \textsuperscript{\rm 3}Artificial Intelligence Research Institute, Shenzhen MSU-BIT University\\
    \{dengqi.kei; niushuaicheng; zhangronghao16; chenyaofo; runhaozeng.cs\}@gmail.com; ellachen@scut.edu.cn
    
}

\usepackage{bibentry}
\begin{document}

\maketitle

\begin{abstract}
Test-time adaptation (TTA) aims to fine-tune a trained model online using unlabeled testing data to adapt to new environments or out-of-distribution data, demonstrating broad application potential in real-world scenarios. However, in this optimization process, unsupervised learning objectives like entropy minimization frequently encounter noisy learning signals. These signals produce unreliable gradients, which hinder the model’s ability to converge to an optimal solution quickly and introduce significant instability into the optimization process. In this paper, we seek to resolve these issues from the perspective of optimizer design. Unlike prior TTA using manually designed optimizers like SGD, we employ a learning-to-optimize approach to automatically learn an optimizer, called Meta Gradient Generator (MGG). Specifically, we aim for MGG to effectively utilize historical gradient information during the online optimization process to optimize the current model. To this end, in MGG, we design a lightweight and efficient sequence modeling layer -- gradient memory layer. It exploits a self-supervised reconstruction loss to compress historical gradient information into network parameters, thereby enabling better memorization ability over a long-term adaptation process. We only need a small number of unlabeled samples to pre-train MGG, and then the trained MGG can be deployed to process unseen samples. Promising results on ImageNet-C/R/Sketch/A indicate that our method surpasses current state-of-the-art methods with fewer updates, less data, and significantly shorter adaptation times. Compared with a previous SOTA SAR, we achieve 7.4\% accuracy improvement and 4.2$\times$ faster adaptation speed on ImageNet-C. Code: https://github.com/keikeiqi/MGTTA.
\end{abstract}

\section{Introduction}
\label{sec:intro}
Since the emergence of test-time adaptation (TTA) \cite{TTT,TENT}, it has made significant progress~\cite{TTT++,MEMO,T3A,DUA,LAME,gandelsman,foa} and demonstrated broad potential across various scenarios to enhance model performance on out-of-distribution (OOD) data or novel environments, also known as distribution shifts. By adapting to each test data immediately after inference in an unsupervised manner, TTA offers minimal overhead, distinguishing it as a practical choice for real-world applications.

Although the online unsupervised setting enhances TTA's practicality, it also introduces certain challenges for TTA.
This is because the model performance often degrades significantly on OOD data, leading to unsupervised objectives like entropy minimization \cite{TENT} and self-learning \cite{goyal2022test} encountering noisy supervision, which in turn produces unreliable gradients. Such unreliable gradients may hinder TTA from converging to an optimal solution quickly and also introduce instability into the learning process. This issue is especially critical in more complex test settings, such as label distribution shifts and mixed domain shifts, as highlighted by \citet{SAR}.

To address this issue, various TTA methods \cite{EATA,SAR,yuan2023robust,deyo} have been developed. For example, both EATA \cite{EATA} and DeYO \cite{deyo} devise different sample filtering strategies to select partial samples for TTA. However, the sample filtering strategy sometimes might be threshold-sensitive, making it difficult to set a reasonable threshold when test data are unknown. Moreover, filtering samples may result in the loss of valuable information, leading to insufficient learning, particularly when only a few test samples are available. In addition to sample filtering, EATA \cite{EATA} also exploits weight regularization and SAR \cite{SAR} devises a sharpness-aware minimization strategy to stabilize the online TTA process, \etc~

\begin{figure}[!t]
    \centering 
    \includegraphics[width=0.85\linewidth]{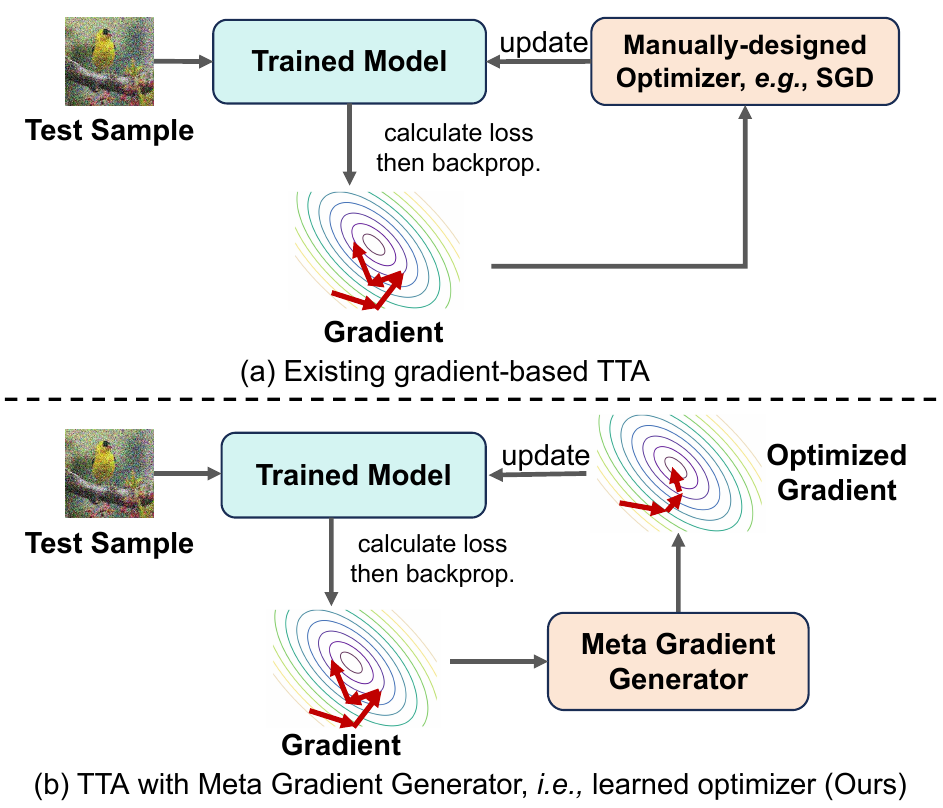}
    \vspace{-0.05in}
    \caption{Method Differences. We devise an automatically learned meta gradient generator to optimize the original gradients produced by a TTA loss to be more reliable.} 
    \label{fig1}
    \vspace{-0.15in}
\end{figure}

Unlike prior methods, we seek to resolve the above issue from a new perspective of optimizer design. Instead of using manually designed optimizers like SGD and Adam, we cast the design of optimization algorithms as a learning problem, \ie, learning to optimize (L2O)~\cite{andrychowicz2016learning}. \rh{Benefiting from the strong power of end-to-end learning, L2O has been extensively validated in supervised settings, even with noisy input gradients, demonstrating its ability to enhance model performance and accelerate convergence. Inspired by this, we devise an automatically learned optimizer for TTA, as shown in Figure~\ref{fig1}.} 
Considering that in online TTA, all unreliable gradients are immediately used for model updates after inference, their cumulative effect can amplify the impact of these unreliable gradients. \rh{However, noisy gradients are often short-term fluctuations, while the model optimization/gradients typically exhibit regularity over a longer time scale.} In this paper, we suggest that the issue of unreliable gradients indeed can be alleviated a lot if we have access to all test sample gradients before adaptation, \rh{as leveraging the patterns in the historical gradient path would allow for a collective analysis to determine a more reliable gradient descent direction. This shares a similar motivation with SGDM or Adam that exploit historical gradients to improve the optimization.}
Thus, we propose memorizing \rh{long-term} historical gradients during online adaptation to help and guide L2O in generating more reliable gradients.

Based on the above motivation, we devise a \dq{\textbf{M}eta \textbf{G}radient Generator (MGG)-guided \textbf{T}est-\textbf{T}ime \textbf{A}daptation method, termed (\textbf{MGTTA})}, in which the MGG is \rh{built upon L2O to replace manually designed optimizers used in TTA.} MGG first memorizes the input original gradients and then outputs the refined gradients for TTA. For memorization, we devise a lightweight and efficient sequence modelling layer, termed the Gradient Memory Layer (GML). GML is inspired by a most recent advanced architecture in the large language model community, termed \textbf{test-time training} layer \cite{sun2024learning}. GML leverages a self-supervised reconstruction loss to encode historical gradient information into the model parameters, thereby enhancing GML’s capacity to retain all historical gradients throughout a long-term online adaptation process. For gradient optimization/correction, we utilize a feature discrepancy loss and a prediction entropy loss as our TTA objectives, guiding the MGG in automatically refining the original gradients to enhance their reliability through a learning-to-optimize manner. The training of MGG only requires a small number of unlabeled OOD samples (\eg, 128), and then the trained MGG can be deployed to unseen samples for TTA. 
Extensive results indicate that our method surpasses existing SOTAs with fewer updates, fewer data, and significantly shorter adaptation times.

    We summarize our main novelty and contributions below. 
    
    $\bullet$ We devise a novel Meta Gradient Generator (MGG), which is automatically learned in a learning-to-optimize manner, to replace manually designed optimizers for TTA. This generator takes the original unreliable gradients as input and produces optimized gradients, thereby alleviating the impact of noisy learning signals encountered in TTA.
    
    $\bullet$ We introduce a lightweight yet efficient sequential modeling network, namely Gradient Memory Layer, which can memorize the historical gradient information during a long-term online TTA process by encoding this information into model parameters via a reconstruction loss. Then, we use the memorized gradients to guide the optimization/correction of gradients for the current model adaptation.
    
    $\bullet$ Extensive experiments demonstrate that using a small number (\eg, 128) of unlabeled samples from the ImageNet-C validation set is sufficient to train an effective MGG. With this pre-trained MGG, our method outperforms existing methods on various unseen datasets, including ImageNet-C/R/Sketch/A. The fast convergence observed in experiments makes our method practical for real-world scenarios especially when the computational resource budget is limited, as shown in Table~\ref{limit_time_samples}.

\section{Related Work}
\label{sec:related_work}

\textbf{Test-Time Adaptation (TTA)} aims to adapt a trained model to new environments or OOD data online using unlabeled test data and it has made considerable progress~\cite{nado2020evaluating, khurana2021sita, boudiaf2022parameter, AME, chencola, cema, wen2023test}. According to whether relying on gradient computation, existing TTA methods can be categorized into gradient-free methods, such as LAME~\cite{LAME} and T3A~\cite{T3A}, and gradient-based methods, such as TENT~\cite{TENT}, EATA~\cite{eata-c}, \etc~ By directly updating the model parameters using some unsupervised loss, the later one often achieves much better performance. However, in this unsupervised learning, the objectives may generate unreliable gradients due to the interference of noisy signals, leading to unstable adaptation or suboptimal performance. To address this, several methods have been proposed. For example, CoTTA~\cite{cotta} mitigates error accumulation by utilizing weight-averaged and augmentation-averaged pseudo-labels. SAR~\cite{SAR} introduces a sharpness-aware entropy minimization, while both SAR and DeYO~\cite{deyo} devise sample filtering strategies to exclude certain samples from adaptation. However, these methods may sometimes be threshold-sensitive or require substantial additional computational resources. 
Unlike existing methods, in this paper, we explore a new perspective for resolving the unreliable gradients issue, \ie, exploiting a learning-to-optimize framework to automatically learn a gradient optimizer for TTA, thereby enhancing the quality of gradients used for TTA.

\begin{figure*}[!t]
    \centering 
    \includegraphics[width=\linewidth]{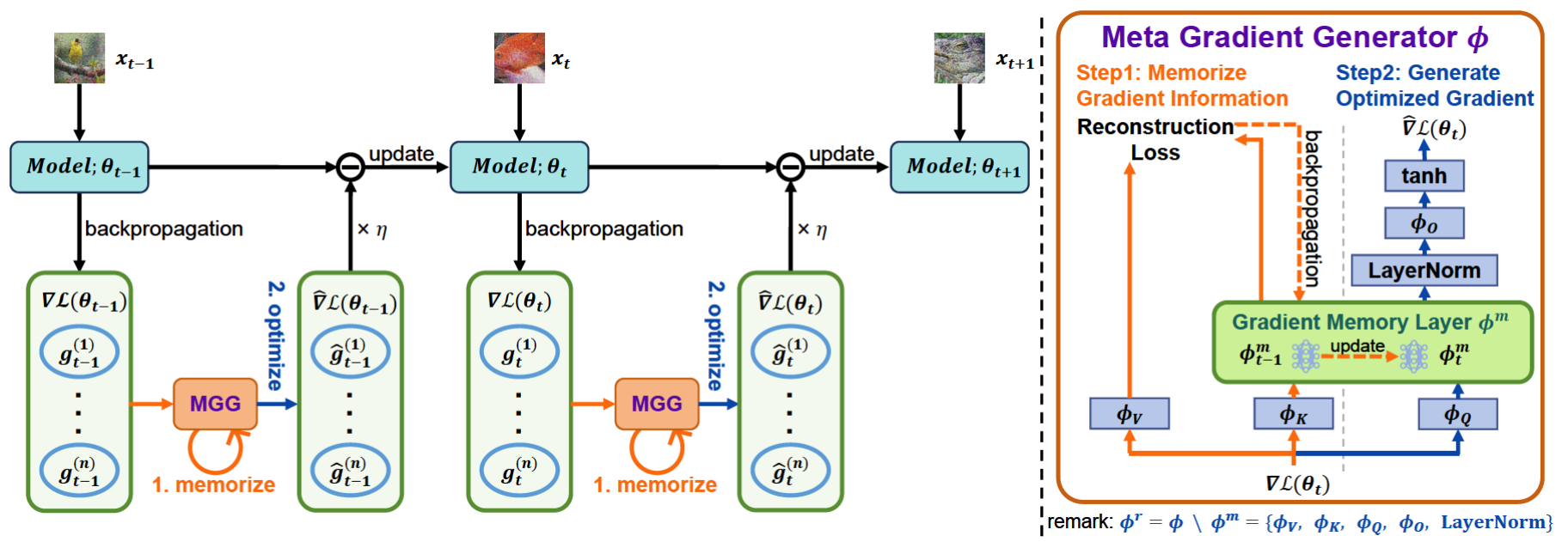}
    \vspace{-0.2in}
    \caption{An overall illustration of \methodname, in which we design a two-step meta gradient generator (MGG) to generate optimized gradients for TTA. Given a trained model $f(\cdot;\theta)$, for each batch of test samples, we first calculate predictions and obtain gradients by backpropagation. Then, in Step 1  MGG first memorizes gradients and then in Step 2 MGG generates optimized gradients based on the historical gradient information. Finally, the model parameters \( \theta \) are updated using the optimized gradients. Here, the learnable parameters within $\theta$ only involve norm layers and the rest are kept frozen during adaptation.}
    \label{fig:overall}
\end{figure*}

\noindent\textbf{Learning to Optimize (L2O)} aims to automatically learn optimizers \cite{li2016learning}, improving on traditional methods like Bayesian optimization, random search, and gradient-based approaches.  
\citet{andrychowicz2016learning} treated optimization as a learning problem, and subsequent methods extended it by introducing techniques such as task-independent optimization~\cite{li2017learning}, hierarchical RNN architectures~\cite{wichrowska2017learned}, and random scaling to speed up training~\cite{lv2017learning}. To address gradient truncation, solutions like dynamic weighting~\cite{metz2019understanding} and progressive unroll length~\cite{chen2020training} were proposed. HALO~\cite{li2020halo} enhanced generalization through a novel regularizer, while PES~\cite{vicol2021unbiased} removed truncation bias using accumulated correction terms. SL2O~\cite{chen2022scalable} focused on learning within a restricted subspace. The above methods show promise in speeding up model convergence and highlight the potential of L2O approaches. However, they rely on ground truth for training, which is often unavailable in real-world scenarios. In this paper, we explore the concept of L2O to optimize gradients during the TTA process and design a self-supervised training method that enables efficient L2O without the need for labeled data.

\section{Proposed Method}
\label{sec:proposed_method}

\subsection{Problem Statement}
Consider a set of source training images $\{x_n\}_{n=1}^N$, where each image $x_n$ is drawn from the distribution $P(x)$. A model $f(\cdot;\theta)$, parameterized by \( \theta \), is trained using the labeled dataset $\{(x_n, y_n)\}_{n=1}^N$. Ideally, $f(\cdot;\theta)$ performs effectively on test samples originating from the same distribution as the training data, denoted by $x\sim P(x)$. However, in practice, this condition is frequently unmet. The test data may often be out of distribution, possibly corrupted, represented by $x\sim U(x)$, where $U(x) \neq P(x)$. Such discrepancies can significantly impair the model's performance.

Test-time adaptation (TTA) attempts to mitigate this issue by adapting the model using only unlabeled test samples.
During the TTA phase, the model $f(\cdot;\theta)$ often employs an unsupervised loss $\mathcal{L}(\theta)$ to fine-tune its parameters, thus facilitating adaptation to out-of-distribution (OOD) data or new environments. Traditional TTA methods typically use manually designed optimizers like stochastic gradient descent (SGD) to update the parameters at the step $t\small{+}1$ by 
\begin{eqnarray}
\theta_{t+1} = \theta_t - \eta \cdot \nabla \mathcal{L} (\theta_t),
\label{eqn:gd}
\end{eqnarray}
where $\eta$ denotes the learning rate.
However, such unsupervised learning objectives may produce unreliable gradients due to noise interference, resulting in unstable optimization and challenges in swiftly converging to an optimal solution.

\subsection{General Scheme}
To address the above issues, we propose a learning-to-learn framework to develop a novel optimizer tailored for TTA scenarios. We learn to generate gradients by developing a neural network-based optimizer termed the \textbf{meta gradient generator} $f_{\textrm{MGG}}(\cdot;\phi)$, where $\phi$ is learnable parameters. This optimizer refines the current gradient $\nabla \mathcal{L} (\theta_t)$ to yield more precise and stable gradients and updating parameters by
\begin{eqnarray}\label{eq:tta_update}
\theta_{t+1} = \theta_t - \eta \cdot f_{\textrm{MGG}}(\nabla \mathcal{L} (\theta_t);\phi).
\label{eqn:mgg}
\end{eqnarray}
Here, the key challenge is how can we design MGG to make it optimize the $\nabla \mathcal{L} (\theta_t)$ to be more reliable. In this paper, we posit that unreliable gradients primarily arise due to the online nature of TTA. Since TTA processes each test sample only once for immediate adaptation post-inference, any unreliable gradients shall be incorporated into the model adaptation process, thereby degrading performance. Ideally, if we could access all test samples’ gradients before adaptation, we could analyze them collectively and determine a more reliable direction for gradient descent, thus mitigating the impact of unreliable gradients. However, this approach conflicts with the online nature of TTA, potentially converting it into an offline process. Instead, another straightforward solution is storing gradients from previous test samples during online learning, which, however, will result in much higher memory consumption. To address this, we propose to use MGG to memorize historical gradients during the online adaptation process, and then leverage this historical information to help MGG generate more reliable gradients.

To be specific, we propose a compact yet effective sequential modeling scheme, termed as \textbf{gradient memory layer} \( f_{\textrm{GML}}(\cdot;\phi^m) \), which serves as a pivotal component within our MGG ($\phi^m \small{\in} \phi$). Inspired by Test-Time Training layers~\cite{sun2024learning}, our design idea for \( f_{\textrm{GML}}(\cdot) \) is ``parameter as memory"--compresses the continuous gradient update information into the neural network parameters. Without loss of generality, consider the \( t \)-th step of TTA, our method operates in two steps: \textbf{1) memorize}, input the gradient \( \nabla\mathcal{L} (\theta_t) \), directly computed from the loss, into \( f_{\textrm{GML}} \), and updating the parameters $\phi^m$ of $f_\textrm{{GML}}(\cdot;\phi^m)$ via self-supervised learning to encode the current information. \textbf{2) optimize}, use updated \( \phi^m \) to optimize the input gradient, ultimately applying Eqn. (\ref{eqn:mgg}) for TTA. The schematic depiction of our approach is shown in Figure~\ref{fig:overall}. We term our method as \textbf{M}eta \textbf{G}radient Generator-guided \textbf{T}est-\textbf{T}ime \textbf{A}daptation (\textbf{MGTTA}). Below we first introduce the details of MGG and then the pre-training and TTA pipelines of our MGTTA.

\subsection{Meta Gradient Generator (MGG)}
Unlike prior gradient-based TTA methods that utilize manually designed optimizers, \eg, SGD, we seek to update the model at test time using an auto-learned optimizer, termed the meta gradient generator (MGG). MGG takes the gradients computed from the unsupervised loss as input and outputs optimized gradients for more stable parameter updates. 

\subsubsection{Gradient Memory Layer} 
TTA typically operates online and processes sequential samples, in which each sample (with its gradients) is discarded immediately after adaptation. This process is highly sensitive to unreliable gradients since in this way all unreliable signals will be accumulated and finally degrade model performance a lot. Therefore, we envision the MGG being capable of memorizing historical gradient information. In this sense, MGG shall have the potential to understand the previous and current gradients together, and thus determine a more reliable gradient descent direction, alleviating the impacts of unreliable signals.

To achieve the above goal, one can employ Long Short-Term Memory (LSTM) networks~\cite{lstm} to store the history of time series in a hidden state vector. However, this method is often constrained by the expressive power of a single vector when managing long sequences. Inspired by the Test-Time Training layer~\cite{sun2024learning} that most recently emerged in the large language model community, we introduce a gradient memory layer $f_{\textrm{GML}}(\cdot; \phi^m)$, to compress the sequence information within network parameters, which has much stronger expression power as model parameters have larger capacity than the hidden states of LSTM. \ie, higher dimension. We design a two-step learning approach to implement MGG.

\subsubsection{Step 1: Memorize Gradient Information}
The central concept in incorporating gradient information compression into \( f_{\textrm{GML}} \) is to utilize a reconstruction loss as the self-supervision to update $\phi^m$. This approach is similar to how language models often employ reconstruction or mask prediction loss to compress knowledge from the training corpus into neural network parameters through gradient descent. \dq{During test time, GML is continuously updated by learning from test data, allowing it to “memorize” and leverage historical information in future TTA steps.}
Specifically, suppose the input gradient at time $t$ is $g_t=\nabla \mathcal{L} (\theta_t)$, the reconstruction loss can be expressed as
\begin{eqnarray}\label{eq:reconstruction_loss}
    \mathcal{L}_\textrm{{GML}}(g_t;\phi^m) = \parallel f_\textrm{{GML}}(\phi_K g_t; \phi^m) - \phi_V g_t \parallel^2,
    \label{eqn:gml}
\end{eqnarray}
where $\phi_K$ and $\phi_V$ are two learnable projection matrices that upscale $g_t$, similar but different from Test-Time Training~\cite{sun2024learning} in which the projection matrices are low-rank. 
We then update $\phi^m$ by
\begin{eqnarray}\label{eq:update_w}
\phi^m\leftarrow \phi^m - \eta_\textrm{{GML}} \cdot \nabla \mathcal{L}_\textrm{{GML}}(g_t;\phi^m),
\end{eqnarray}
where $\eta_\textrm{{GML}}$ is the learning rate. Here, since the learning rate is crucial in gradient descent, we use an adaptive learning rate that is determined by a learnable vector $\phi_{lr}$, denoted by
\begin{eqnarray}
    \eta_\textrm{{GML}} = \sigma (\phi_{lr} \cdot g_t),
\end{eqnarray}
where $\sigma$ is the sigmoid function. Up to this time point, $\phi^m$ contains gradient information from before and at time step $t$, achieving the memorization of current gradient information in the network parameters.

\subsubsection{Step 2: Generate Optimized Gradient}
With the updated memory that encapsulates both current and historical gradient information, we can commence the optimization of gradients. The gradient \(g_t\) in need of optimization is first processed through a projection layer \(\phi_Q\) (akin to \(\phi_K\) and \(\phi_V\)), subsequently fed into the updated function \(f_{\textrm{GML}}(\cdot; \phi^m)\), followed by a layer normalization operation and another projection layer \(\phi_O\). The process of obtaining the optimized gradient can be represented as
\begin{eqnarray}\label{eq:correct_gradient}
    \hat{g}_t \small{=} f_{\textrm{MGG}}(g_t;\phi) \small{=} \tanh(\phi_O \cdot \textrm{LN} (f_\textrm{{GML}}(\phi_Q g_t;\phi^m))),
\end{eqnarray}
where $\textrm{LN}$ denotes a LayerNorm layer. Then, this gradient $\hat{g}_t$ can be applied to the TTA of our target model (Eqn. (\ref{eqn:mgg})).

\subsection{Learning MGG before Test-Time Adaptation}
Considering that MGG divides the gradient optimization process into two steps, we accordingly design a multi-step iterative update strategy for the MGG parameters. Without loss of generality, let \(\phi\) be the parameters of MGG, by excluding those of $f_{\textrm{GML}}(\cdot,\phi^m)$, we denote the \textbf{remaining parameters} as \textbf{\(\phi^{r} = \phi \setminus \phi^m\)}.
In other words, all these parameters \(\phi_Q\), \(\phi_K\), \(\phi_V\)and \(\phi_O\) belong to \(\phi^{r}\) (as shown in the right part of Figure~(\ref{fig:overall})).
\textbf{First}, input the data \(x_t\) into the model \(f(\cdot; \theta)\) to obtain predictions $\hat{y}$ (\eg, classification results). Then, use the loss function \(\mathcal{L}_{TTA}(\theta)\) to perform backpropagation to obtain gradients \wrt $\phi^r$ and $\theta$. \textbf{Second}, update $\phi^r$ using the corresponding gradients. \textbf{Third}, fix the parameters \(\phi^{r}\) and use Eqn. (\ref{eqn:gml}) to update the $f_{\textrm{GML}}$ parameters \(\phi^m\). \textbf{Lastly}, with the parameters \(\phi^m\) and \(\phi^r\) fixed, utilize MGG to obtain the optimized gradient and employ Eqn. (\ref{eqn:gd}) to update the parameters \(\theta\) of the model requiring TTA. 

For the TTA loss, we select FOA~\cite{foa} as our test-time learning objective as FOA loss is one of the most recent advanced objectives and has shown promising performance in the existing TTA literature. Beyond entropy minimization~\cite{TENT}, FOA introduces a feature statistics discrepancy loss. To be specific, FOA first collects a small set of unlabelled in-distribution samples (\eg, 64 samples) to calculate the source feature mean $\{\mu_i^S\}_{i=1}^N$ and variances $\{\sigma_i^S\}_{i=1}^N$ of each layer, where $i$ denotes layer index. During TTA, FOA also calculates the corresponding statistics $\{\mu_i(\mathcal{X}_t)\}_{i=1}^N$ and $\{\sigma_i(\mathcal{X}_t)\}_{i=1}^N$ of testing samples and aligns them with pre-calculated source statistics. Formally, given a batch of test samples $\mathcal{X}_t$, the loss is defined by
\begin{flalign}\label{eq:tta_loss}
     \mathcal{L}_{TTA}&(f(\mathcal{X}_t;\theta)) = \sum_{x \in \mathcal{X}} \sum_{c \in \mathcal{C}} -\hat{y}_c \log \hat{y}_c + \nonumber \\ 
     &\lambda \sum_{i=1}^{N} (\parallel \mu_i(\mathcal{X}_t)-\mu_i^S \parallel^2 + \parallel \sigma_i(\mathcal{X}_t) - \sigma_i^S \parallel^2),
\end{flalign}
where $C$ denotes the number of categories, $\hat{y}_c$ is the prediction for category $c$, and $\lambda$ is a trade-off parameter. Note that the entire training process of MGG only requires a few number of unlabeled test samples. In our main experiments, we random sample 128 images without labels from the held-out validation set of ImageNet-C as the training set of MGG, and then test the trained MGG on all ImageNet-C testing datasets and other ImageNet variants. Experiments in Table~\ref{table:train_data_num} demonstrate that 128 images are sufficient for our method to achieve excellent performance.

\subsection{Test-Time Adaptation with MGG}
With a trained MGG, we can begin to employ it for TTA. TTA is an online optimization process with limited available resources. To save computational and storage overhead while efficiently adapting the model, we update only the parameters of the model's normalization layers (\eg, 0.044\% of parameters in ViT-Base~\cite{vit}). 

\subsubsection{Parameter-wise Memorizing}To enhance the capability of MGG in modeling the temporal variations of gradients, we treat each parameter independently rather than as a whole.
Specifically, within TTA, each parameter $\theta^{i}$ within the model parameters $\theta \in \mathbb{R}^n$ evolves over time, forming a sequence $\{\theta_0^{i}, \theta_1^{i}, \ldots, \theta_t^{i}\}$. For each parameter $\theta^i$, we employ an independent $\phi^m$ to model the corresponding sequence of gradient information, while the other parameters of the MGG, \ie, $\phi^r$, are shared across all parameters of $\theta$. This approach allows the GML to concentrate on the temporal changes of each individual parameter rather than the interrelationships among them. Although maintaining historical information for each parameter incurs certain memory costs, the GML is a compact neural network with a small number of parameters, resulting in minimal additional memory overhead during TTA. Taking ViT-base as an example ($n=38,400$), in our implementation, $\phi^m$ for each $\theta^i$ is a linear layer with input and output dimensions of 8, making the total parameters of all GML summing up to only 2.76 M.

\begin{algorithm}[!t]
    \caption{The pre-training/TTA pipeline of MGTTA.\\
    // \emph{Pre-training\&TTA share the same pipeline but differ from:}
     \emph{For \textbf{pre-training}, we use $\mathcal{D}_{val}$ and random initialized $\phi^r \small{\in} \phi$}. \\
    \emph{For \textbf{TTA}, we use $\mathcal{D}_{test}$ and $\phi^r$ is inherited from pre-training.}}

    \begin{algorithmic}[1]\label{alg_1}
        \REQUIRE{Trained model $f(\cdot;\theta)$, MGG model $f_{\textrm{MGG}}(\cdot;\phi)$, samples $\mathcal{D}\small{=}\{x_j\}_{j=1}^M$, hyper-parameters $T$ and $\eta$.}
        \STATE Random initialize GML's parameters as $\phi^m_1$ 
        \STATE Calculate predictions on batch $\mathcal{X}_1$ from $\mathcal{D}$ via $f(\cdot;\theta)$
        \STATE Calculate the loss $\mathcal{L}_{\text{TTA}}(f(\mathcal{X}_1;\theta))$ in Eqn.~(\ref{eq:tta_loss})
        \STATE Calculate the gradient $g_1\small{=}{\nabla}\mathcal{L}_{\text{TTA}}(\theta)$
        \FOR{$t$ in $[2,3,\cdots,T]$}
            \STATE // \emph{\textbf{Step1: Memorize Gradient Information}}
            \STATE Calculate the loss $\mathcal{L}_{\text{GML}}(g_{t-1};\phi^m_{t-1})$ in Eqn.~(\ref{eq:reconstruction_loss})
            \STATE Obtain $\phi^m_t$ with ${\nabla}\mathcal{L}_{\textrm{GML}}(\phi^m_{t-1})$ via Eqn.~(\ref{eq:update_w})
            
            \STATE // \emph{\textbf{Step2: Generate Optimized Gradient for TTA}}
            \STATE Optimize the gradient 
            $\hat{g}_{t-1} \small{=} f_{\textrm{MGG}}(g_{t-1};\phi)$
            
            \STATE Update $\theta$ with optimized gradient $\hat{g}_{t-1}$ via Eqn.~(\ref{eq:tta_update})
            \STATE Calculate predictions on batch $\mathcal{X}_t$ from $\mathcal{D}$ via $f(\cdot;\theta)$

            \STATE Calculate the loss $\mathcal{L}_{\text{TTA}}(f(\mathcal{X}_t;\theta))$ in Eqn.~(\ref{eq:tta_loss})
            \STATE Calculate gradients ${\nabla}\mathcal{L}_{\text{TTA}}(\phi^r)$ and $g_t\small{=}{\nabla}\mathcal{L}_{\text{TTA}}(\theta)$
            \STATE Update $\phi^r$ via ${\nabla}\mathcal{L}_{\text{TTA}}(\phi^r)$ 
        \ENDFOR
        \ENSURE The trained MGG $f_{\textrm{MGG}}(\cdot;\phi)$ for \textbf{pre-training} \textbf{or} \\ The predictions of all samples in $\mathcal{D}$ for\textbf{ TTA}.
    \end{algorithmic}
    \label{alg:overall}
 \end{algorithm}

\subsubsection{Deploying Pre-trained MGG for TTA}
After a small set of held-out samples are used for training, the MGG can be directly employed for TTA. Initially, input the data \(x\) into the model \(f(\cdot; \theta)\) and compute the initial gradient \(\mathcal{L}_{TTA}(\theta)\). Subsequently, input \(g\) into the MGG to perform one memorization and one optimization step, then use the optimized gradient to update the model \(f(\cdot; \theta)\). Note that during TTA, one can choose whether to update $\phi^r$. However, our preliminary studies indicate that this does not clearly improve performance. Therefore, for higher efficiency, we keep $\phi^r$ frozen during the whole TTA phase.
We summarize the overall pseudo-code of our method in algorithm~\ref{alg_1}.

\begin{table*}[!ht]
\setlength{\tabcolsep}{1mm}
\renewcommand\arraystretch{1}
\centering
\fontsize{9}{9}\selectfont
    \begin{tabular}{lcccccccccccccccc}
    \multicolumn{1}{c}{}               & \multicolumn{3}{c}{Noise}                                          & \multicolumn{4}{c}{Blur}                                                           & \multicolumn{4}{c}{Weather}                                                        & \multicolumn{4}{c}{Digital}                                                        &               \\
    \multicolumn{1}{l|}{Method}        & Gauss.        & Shot          & \multicolumn{1}{c|}{Impul.}        & Defoc.        & Glass         & Motion        & \multicolumn{1}{c|}{Zoom}          & Snow          & Frost         & Fog           & \multicolumn{1}{c|}{Brit.}         & Contr.        & Elastic       & Pixel         & \multicolumn{1}{c|}{JPEG}          & Avg.          \\ \midrule
    \multicolumn{1}{l|}{NoAdapt}       & 56.8          & 56.8          & \multicolumn{1}{c|}{57.5}          & 46.9          & 35.6          & 53.1          & \multicolumn{1}{c|}{44.8}          & 62.2          & 62.5          & 65.7          & \multicolumn{1}{c|}{77.7}          & 32.6          & 46.0          & 67.0          & \multicolumn{1}{c|}{67.6}          & 55.5          \\
    
    \multicolumn{1}{l|}{TENT}          & 60.3          & 61.6          & \multicolumn{1}{c|}{61.8}          & 59.2          & 56.5          & 63.5          & \multicolumn{1}{c|}{59.2}          & 54.3          & 64.5          & 2.3           & \multicolumn{1}{c|}{79.1}          & 67.4          & 61.5          & 72.5          & \multicolumn{1}{c|}{70.6}          & 59.6          \\
    \multicolumn{1}{l|}{CoTTA}         & 63.6          & 63.8          & \multicolumn{1}{c|}{64.1}          & 55.5          & 51.1          & 63.6          & \multicolumn{1}{c|}{55.5}          & 70.0          & 69.4          & 71.5          & \multicolumn{1}{c|}{78.5}          & 9.7           & 64.5          & 73.4          & \multicolumn{1}{c|}{71.2}          & 61.7          \\
    \multicolumn{1}{l|}{SAR}           & 59.2          & 60.5          & \multicolumn{1}{c|}{60.7}          & 57.5          & 55.6          & 61.8          & \multicolumn{1}{c|}{57.6}          & 65.9          & 63.5          & 69.1          & \multicolumn{1}{c|}{78.7}          & 45.7          & 62.4          & 71.9          & \multicolumn{1}{c|}{70.3}          & 62.7          \\
    \multicolumn{1}{l|}{FOA}           & 61.5          & 63.2          & \multicolumn{1}{c|}{63.3}          & 59.3          & 56.7          & 61.4          & \multicolumn{1}{c|}{57.7}          & 69.4          & 69.6          & 73.4          & \multicolumn{1}{c|}{81.1}          & 67.7          & 62.7          & 73.9          & \multicolumn{1}{c|}{73.0}          & 66.3          \\
    \multicolumn{1}{l|}{EATA}          & 61.2          & 62.3          & \multicolumn{1}{c|}{62.7}          & 60.0          & 59.2          & 64.7          & \multicolumn{1}{c|}{61.7}          & 69.0          & 66.6          & 71.8          & \multicolumn{1}{c|}{79.7}          & 66.8          & 65.0          & 74.2          & \multicolumn{1}{c|}{72.3}          & 66.5          \\
    \multicolumn{1}{l|}{DeYO}          & 62.4          & 64.0          & \multicolumn{1}{c|}{63.9}          & 61.0          & 60.7          & 66.4          & \multicolumn{1}{c|}{62.9}          & 70.9          & 69.6          & 73.7          & \multicolumn{1}{c|}{80.5}          & 67.2          & 69.9          & 75.7          & \multicolumn{1}{c|}{73.7}          & 68.2          \\ \midrule

    \multicolumn{1}{l|}{\mgtta (ours)} & \textbf{64.5} & \textbf{66.5} & \multicolumn{1}{c|}{\textbf{66.3}} & \textbf{63.8} & \textbf{65.0} & \textbf{70.1}          & \multicolumn{1}{c|}{\textbf{69.7}}          & \textbf{74.5} & \textbf{72.8} & \textbf{77.0} & \multicolumn{1}{c|}{\textbf{81.3}}          & \textbf{71.0} & \textbf{75.0} & \textbf{77.7} & \multicolumn{1}{c|}{\textbf{75.1}} & \textbf{71.3}
    \end{tabular}
\vspace{-0.2cm}
\caption{Comparisons with state-of-the-art methods on ImageNet-C \wrt accuracy(\%).}
\label{ImageNet-C}
\end{table*}

\begin{table*}[t]
\setlength{\tabcolsep}{1.8mm}
\renewcommand\arraystretch{1}
\centering
\fontsize{9}{9}\selectfont
    \begin{tabular}{l|cccccccc|cccccccc}
    \multicolumn{1}{c}{}      & \multicolumn{8}{c}{Time Budget for Adaptation (seconds)}                                                                                   & \multicolumn{8}{c}{\# Data Budget for Adaptation (number of batch)}                                                                                   \\ 
    Method                        & 2.0           & 5.0           & 10.0          & 20.0          & 30.0          & 50.0          & 90.0          & $\infty$             & 10            & 20            & 35            & 50            & 75            & 100           & 200           & 782           \\ \midrule
    NoAdapt                 & 55.5          & 55.5          & 55.5          & 55.5          & 55.5          & 55.5          & 55.5          & 55.5          & 55.5          & 55.5          & 55.5          & 55.5          & 55.5          & 55.5          & 55.5          & 55.5          \\
    TENT                    & 56.8          & 55.4          & 56.3          & 57.8          & 58.7          & 59.4          & 59.8          & 59.6          & 56.7          & 55.7          & 55.4          & 56.0          & 56.7          & 57.4          & 58.8          & 59.6          \\
    CoTTA                   & 45.8          & 37.5          & 37.1          & 36.3          & 37.7          & 38.3          & 39.5          & 61.7          & 36.6          & 36.8          & 38.4          & 37.3          & 38.1          & 39.6          & 42.6          & 61.7          \\
    SAR                     & 55.9          & 56.6          & 56.4          & 55.8          & 56.4          & 61.3          & 58.9          & 62.7          & 56.9          & 57.0          & 52.1          & 55.8          & 56.6          & 57.0          & 62.1          & 62.7          \\
    FOA                     & 54.3          & 55.3          & 56.1          & 56.7          & 57.1          & 57.7          & 59.1          & 66.3          & 56.9          & 57.3          & 58.3          & 59.5          & 61.1          & 62.1          & 64.2          & 66.3          \\
    EATA                    & 57.0          & 58.8          & 61.5          & 63.9          & 64.8          & 65.8          & 66.4          & 66.5          & 56.7          & 58.1          & 59.6          & 60.6          & 62.1          & 63.1          & 65.0          & 66.5          \\
    DeYO                    & 57.0          & 59.2          & 62.4          & 64.8          & 66.0          & 67.1          & 68.0          & 68.2          & 57.1          & 59.1          & 60.9          & 61.9          & 63.8          & 64.9          & 66.9          & 68.2          \\ \midrule
    \mgtta (ours)                    & \textbf{64.1} & \textbf{68.7} & \textbf{70.4} & \textbf{70.9} & \textbf{71.1} & \textbf{71.3} & \textbf{71.3} & \textbf{71.3} & \textbf{63.2} & \textbf{66.4} & \textbf{69.3} & \textbf{70.1} & \textbf{70.6} & \textbf{70.8} & \textbf{71.2} & \textbf{71.3}
    \end{tabular}
\vspace{-0.2cm}
\caption{Comparisons under limited adaptation budgets on ImageNet-C \wrt acc(\%). Total \#batches is 782, with batch size 64.}
\vspace{-0.05in}
\label{limit_time_samples}
\end{table*}

\begin{table}[t]
\setlength{\tabcolsep}{0.8mm}
\renewcommand\arraystretch{1}
\centering
\fontsize{9}{9}\selectfont
    \begin{tabular}{l|cccccc|c}
    Method & NoAdapt & TENT & SAR  & FOA  & EATA & DeYO & \mgtta          \\ \midrule
    R      & 59.5    & 63.9 & 63.3 & 63.8 & 63.3 & 66.1 & \textbf{70.2} \\
    Sketch & 44.9    & 49.1 & 48.7 & 49.9 & 50.9 & 52.2 & \textbf{53.3} \\
    A      & 0.1     & 52.9 & 52.5 & 51.5 & 53.4 & 54.1 & \textbf{56.7} \\ \midrule
    Avg.   & 34.8    & 55.3 & 54.8 & 55.1 & 55.9 & 57.5 & \textbf{60.1}
    \end{tabular}

\vspace{-0.05in}
\caption{Comparisons with state-of-the-arts on ImageNet-R, ImageNet-Sketch and ImageNet-A \wrt accuracy(\%).}
\vspace{-0.15in}
\label{table:ImageNet-R,Sketch,A}
\end{table}

\section{Experiments}
\label{sec:experiments}
\subsection{Datasets, Models and Compared Methods}
We conduct experiments on four benchmark datasets, including 1) \textbf{ImageNet-C}~\cite{hendrycks2019benchmarking} contains corrupted images in 15 types of 4 main categories and each type has 5 severity levels. \textbf{In ours experiments, all results are evaluated on the severity level 5.} 2) \textbf{ImageNet-R}~\cite{ImageNet-R} contains various artistic renditions of 200 ImageNet classes. 3) \textbf{ImageNet-Sketch}~\cite{ImageNet-sketch} includes sketch-style images representing 1,000 ImageNet classes. 4) \textbf{ImageNet-A}~\cite{ImageNet-A} consists of natural adversarial examples. We use ViT-Base~\cite{vit} as the source model, which is trained on ImageNet-1K and adapted to the above datasets.

\textit{Baselines:} LAME~\cite{LAME}, T3A~\cite{T3A}, TENT~\cite{TENT}, CoTTA~\cite{cotta}, SAR~\cite{SAR}, FOA~\cite{foa}, EATA~\cite{EATA}, DeYO~\cite{deyo}.

\subsection{Implementation Details}
\textbf{For pre-training MGG}, we randomly select 128 unlabeled samples from the ImageNet-C validation set. The learning rate is set to 1e-4 for $\theta$ and 1e-2 for $\phi^r$. We update $\theta$ and $\phi^r$ for $T$=2,000 iterations with a batch size of 2. The GML hidden size is set to 8. \textbf{During TTA}, the batch size is 64, and $\phi^r$ is fixed, the learning rate for $\theta$ is set to 1e-3.

\subsection{Comparisons with State-of-the-arts}
\subsubsection{TTA Results on ImageNet-C.}
We report the results of 15 different corruptions on the ImageNet-C dataset (severity level 5) in Table~\ref{ImageNet-C}. The experimental results indicate that our method significantly outperforms other approaches across all corruptions. Compared to T3A, TENT, and CoTTA, our method achieves an average performance improvement of approximately 10\% on ImageNet-C. When compared to the previous best-performing method, DeYO, our method still shows an improvement of 3.1\% (68.2\% vs. 71.3\%), demonstrating the effectiveness of our approach.

\subsubsection{Performance under Limited Adaptation Budget on ImageNet-C.}
In practical applications, the number of samples available for adaptation and the time allotted are often limited. We conduct two experiments: \textbf{1) limited update time}: set a maximum update time $t_{max}$, beyond which the model is frozen and subsequent data are only used for inference. Results in Table~\ref{limit_time_samples} indicate that our method significantly enhances performance with only limited time used for adaptive updates, achieving a 68.7\% accuracy in just 5 seconds, surpassing other methods over their entire update duration.
\textbf{2) limited available samples}: perform TTA using only the first $u_{max}$ batches of samples. From Table~\ref{limit_time_samples}, our method, adapting with only 35 out of 782 batches (4.5\%), achieves a 69.3\% accuracy, already significantly surpassing other methods. In contrast, methods like SAR, EATA, and DeYO, which involve filtering samples and do not utilize the entire data within a batch, show lower efficiency. MGTTA does not filter samples but instead optimizes the gradient using MGG, offering superior effectiveness and efficiency.

\begin{table}[t]
\setlength{\tabcolsep}{1.6mm}
\renewcommand\arraystretch{1}
\centering
\fontsize{9}{9}\selectfont
    \begin{tabular}{lcccc}
    Variants of our method       & C    & R    & Sketch & A     \\ \midrule
    Ours      & \textbf{71.3} & \textbf{70.2} & \textbf{53.3}   & 56.7  \\
    Ours ~~~\textit{replace GML with LSTM}    & 71.3 & 69.4 & 50.3   & \textbf{56.8}  \\
    Ours ~~~\textit{remove MGG} & 70.0 & 67.2 & 51.9   & 55.1 \\
    \end{tabular}
\caption{Effect of MGG and GML on ImageNet-C/R/S/A.}
\vspace{-0.1in}
\label{table:GML_MGG_ablation}
\end{table}

\begin{table*}[th!]
\setlength{\tabcolsep}{1.2mm}
\renewcommand\arraystretch{1}
\centering
\fontsize{9}{9}\selectfont
    \begin{tabular}{l|ccccccccc|cccc}
    \multirow{2}{*}{} & \multirow{2}{*}{NoAdapt} & \multirow{2}{*}{LAME} & \multirow{2}{*}{T3A} & \multirow{2}{*}{TENT} & \multirow{2}{*}{CoTTA} & \multirow{2}{*}{SAR} & \multirow{2}{*}{FOA} & \multirow{2}{*}{EATA} & \multirow{2}{*}{DeYO} & Ours    & Ours    & Ours    & \multirow{2}{*}{Ours}    \\
                      &                          &                       &                      &                       &                        &                      &                      &                       &                       & ($u_{max}$=20) & ($u_{max}$=50) & ($u_{max}$=100) & \\ \midrule
    Acc.(\%)              & 55.5                     & 54.1                  & 56.9                 & 59.6                  & 61.7                   & 62.7                 & 66.3                 & 66.5                  & 68.2                  & 66.4    & 70.1    & 70.8    & 71.3    \\
    Runtime(s)        & 54.3                     & 54.8                  & 125.5                & 122.6                 & 619.3                  & 242.7                & 1636.7               & 127.4                 & 172.4                 & 55.8    & 58.5    & 63.2    & 125.5  
    \end{tabular}
\vspace{-0.1cm}
\caption{
Wall-clock runtime for processing 50,000 images of ImageNet-C on a RTX 4090 GPU, and Acc. averaged over 15 corruptions. $u_{max}$: adapt the model with only the first $u_{max}$ batches. If $u_{max}$ is not specified, all samples are used for adaptation.}
\vspace{-0.1cm}
\label{runtime}
\end{table*}

\begin{table}[th!]
\setlength{\tabcolsep}{1.5mm}
\renewcommand\arraystretch{1}
\centering
\fontsize{9}{9}\selectfont
    \begin{tabular}{lcccc}
    Dataset                         & \textit{N}=64     & \textit{N}=128   & \textit{N}=256   & \textit{N}=1000  \\ \midrule
    ImageNet-C              & 71.3   & 71.3  & 71.2  & 71.3  \\
    ImageNet-R               & 70.0   & 70.2  & 70.1  & 70.1 \\
    \end{tabular}

\caption{Effect on numbers of training samples for MGG \wrt average accuracy over 15 corruptions on ImageNet-C.}
\label{table:train_data_num}
\end{table}

\begin{table}[th!]
\setlength{\tabcolsep}{0.8mm}
\renewcommand\arraystretch{1}
\centering
\fontsize{9}{9}\selectfont

    \begin{tabular}{l|ccccccc|c}
    \multicolumn{1}{c|}{} & NoAdapt & TENT & CoTTA & SAR  & FOA  & EATA & DeYO & Ours          \\ \midrule
    Acc.                  & 53.8    & 52.1 & 46.3  & 60.7 & 61.7 & 64.6 & 64.0 & \textbf{65.9}
    \end{tabular}
\caption{Comparisons \wrt accuracy on ImageNet-C under mixed domain shifts, \ie, the mixture of 15 corruptions.}
\label{table:mix}
\end{table}

\begin{table}[th!]
\setlength{\tabcolsep}{1.5mm}
\renewcommand\arraystretch{1}
\centering
\fontsize{9}{9}\selectfont

    \begin{tabular}{ccccccc}
    Batch size & TENT & SAR  & FOA  & EATA & DeYO & Ours          \\ \midrule
    2          & 57.4 & 64.3 & 66.5 & 67.0 & 63.6 & \textbf{70.3} \\
    4          & 59.2 & 63.8 & 66.6 & 67.2 & 65.5 & \textbf{70.9}
    \end{tabular}

\caption{Comparisons with small batch size on ImageNet-C \wrt average accuracy(\%) over 15 corruptions. }
\label{table:small_bs}
\vspace{-0.1in}
\end{table}

\subsubsection{TTA Results on ImageNet-R/Sketch/A.}
We further explore the generalization ability of the MGG trained on ImageNet-C validation set to datasets with totally different distribution shift types, including ImageNet-R/Sketch/A datasets. The results, shown in Table \ref{table:ImageNet-R,Sketch,A}, indicate that our method outperforms others on these three challenging ImageNet variant datasets by a large margin. Compared to the previous best-performing method, our method achieves an average performance improvement of 2.6\% (57.5\% vs. 60.1\%). This highlights the strong generalization capability of MGG, making it effective on various OOD datasets.

\subsection{Ablation Studies}

\subsubsection{Effectiveness of MGG and GML.}
We conduct experiments by removing MGG or replacing GML with LSTM. From Table \ref{table:GML_MGG_ablation}, without MGG's gradient optimization, accuracy significantly decreases on ImageNet-C/R/Sketch/A, indicating that MGG generates more reliable gradients. 
On ImageNet-C/A, both our GML and LSTM achieve better performance than without MGG, demonstrating that our idea of enhancing TTA performance through MGG-optimized gradient is general and effective. Notably, when using LSTM, the performance on ImageNet-R/Sketch is lower than GML (50.3\% vs. 53.3\%), possibly because MGG was trained on ImageNet-C val data, which visually differs significantly from ImageNet-R/Sketch. Our GML exhibits better generalization performance compared to LSTM.

\subsubsection{Convergence Comparison.}
\begin{figure}[!t]
    \centering 
    \includegraphics[width=0.75\linewidth]{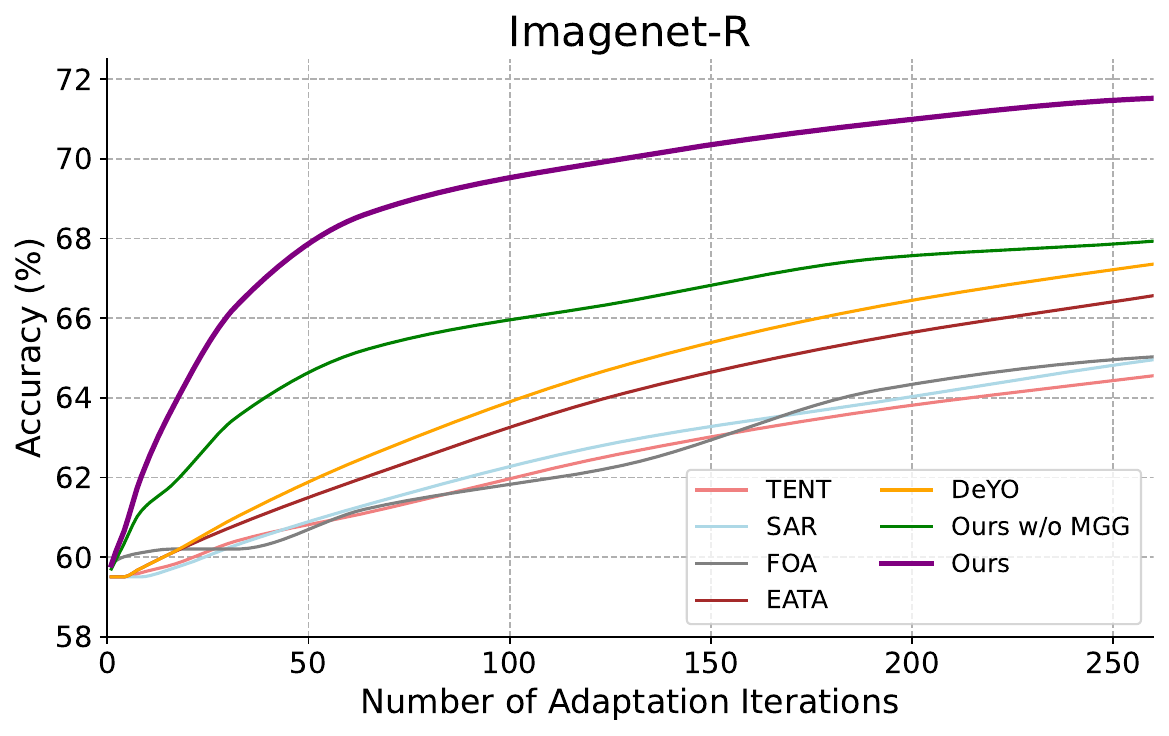}
    \caption{Convergence speed comparisons on ImageNet-R.} 
    \label{converge}
    \vspace{-0.5cm}
\end{figure}

We perform TTA with a fixed number of iterations, followed by accuracy evaluations on remaining samples without additional adaptation. Figure~\ref{converge} shows that \mgtta achieves the best convergence and final accuracy, with a steep rise in accuracy within the initial 0-50 updates, suggesting the effectiveness of MGG-optimized gradients. Here, \mgtta without MGG also achieves better performance than baselines, which mainly benefits from the advanced loss in Eqn.~(\ref{eq:tta_loss}), Even though, our method with MGG still achieves much better convergence than that without MGG, showing the superiority of MGG.

\subsubsection{Effect of the Number of Samples for Pre-training MGG.}
Considering the challenge of data acquisition in practical applications, we explore the feasibility of training MGG with as minimal data as possible. We train MGG with $N$ images randomly sampled from ImageNet-C validation set. From Table \ref{table:train_data_num}, using only 128 samples suffices to achieve optimal accuracies of 71.3\%/70.2\% on ImageNet-C/-R, suggesting our method is able to learn optimizing gradients efficiently.

\subsubsection{Results under Wild Scenarios.} We consider two wild scenarios on ImageNet-C following~\cite{SAR}: \textbf{1)} small TTA batch size and \textbf{2)} mixed distribution shift: mix and randomly shuffle the 15 corruption types. Results in Tables~\ref{table:mix} and~\ref{table:small_bs} show that our method consistently outperforms other methods. Reducing the batch size has a minimal impact on the performance of our method (compared to other methods), indicating that our approach exhibits superior and more stable performance under challenging conditions.

\subsubsection{Runtime and Memory Comparison.} 
\dq{We report the wall-clock time of \mgtta and its variants in Table~\ref{runtime}, in which the \mgtta variants adapt the model using only the first $u_{max}$ out of 782 batches (with batch size 64), and then the model is frozen for the subsequent inference. From the results, \mgtta ($u_{max}=50$) surpasses all baselines (adapt on all 782 batches) by adapting the model on only 50 batches, suggesting our effectiveness. Moreover, \mgtta requires 5,193 MB of GPU memory for TTA, whereas without MGG (\ie, directly updating model parameters via backpropagated gradients) requires 5,165 MB (TENT). Thus, our method consumes only an additional 28 MB of GPU memory, demonstrating that MGG is lightweight and efficient.}

\section{Conclusion}
In this paper, we have proposed a Meta Gradient Generator-guided TTA (MGTTA) method, to alleviate the issues of unreliable gradients encountered during the unsupervised online TTA process. Specifically, we developed a novel neural network-based optimizer to replace manually tuned optimizers. This optimizer has been capable of optimizing gradients, thereby enhancing their reliability and stability. Recognizing that historical gradient information can guide the generation of current gradients, we designed a lightweight and efficient sequence modeling layer, termed gradient memory layer, to store historical information and provide a better direction for gradient optimization. We trained this optimizer using a learning-to-optimize approach with a TTA loss of feature discrepancy and prediction entropy. The superior performance and fast convergence observed in extensive experiments facilitate our application in the real world.

\section*{Acknowledgments}
This work was supported by the National Natural Science Foundation of China (NSFC) (Grant Nos. 62202311, 62376099), Excellent Science and Technology Creative Talent Training Program of Shenzhen Municipality (Grant No. RCBS20221008093224017), Guangdong Basic and Applied Basic Research Foundation (Grant No. 2023A1515011512), Key Scientific Research Project of the Department of Education of Guangdong Province (Grant No. 2024ZDZX3012), Natural Science Foundation of Guangdong Province (Grant No. 2024A1515010989), and Ministry of Education, Singapore, under its Academic Research Fund Tier 1.

\bibliography{aaai25}

\begin{thebibliography}{41}
\providecommand{\natexlab}[1]{#1}

\bibitem[{Andrychowicz et~al.(2016)Andrychowicz, Denil, Gomez, Hoffman, Pfau, Schaul, Shillingford, and De~Freitas}]{andrychowicz2016learning}
Andrychowicz, M.; Denil, M.; Gomez, S.; Hoffman, M.~W.; Pfau, D.; Schaul, T.; Shillingford, B.; and De~Freitas, N. 2016.
\newblock Learning to learn by gradient descent by gradient descent.
\newblock In \emph{Advances in Neural Information Processing Systems}, 3981--3989.

\bibitem[{Boudiaf et~al.(2022{\natexlab{a}})Boudiaf, Mueller, Ben~Ayed, and Bertinetto}]{LAME}
Boudiaf, M.; Mueller, R.; Ben~Ayed, I.; and Bertinetto, L. 2022{\natexlab{a}}.
\newblock Parameter-free online test-time adaptation.
\newblock In \emph{Proceedings of the IEEE Conference on Computer Vision and Pattern Recognition}, 8344--8353.

\bibitem[{Boudiaf et~al.(2022{\natexlab{b}})Boudiaf, Mueller, Ben~Ayed, and Bertinetto}]{boudiaf2022parameter}
Boudiaf, M.; Mueller, R.; Ben~Ayed, I.; and Bertinetto, L. 2022{\natexlab{b}}.
\newblock Parameter-free online test-time adaptation.
\newblock In \emph{Proceedings of the IEEE Conference on Computer Vision and Pattern Recognition}, 8344--8353.

\bibitem[{Chen et~al.(2024{\natexlab{a}})Chen, Niu, Chen, Zhang, Li, Li, and Tan}]{chencola}
Chen, G.; Niu, S.; Chen, D.; Zhang, S.; Li, C.; Li, Y.; and Tan, M. 2024{\natexlab{a}}.
\newblock Cross-device collaborative test-time adaptation.
\newblock In \emph{Advances in Neural Information Processing Systems}.

\bibitem[{Chen et~al.(2020)Chen, Zhang, Jingyang, Chang, Liu, Amini, and Wang}]{chen2020training}
Chen, T.; Zhang, W.; Jingyang, Z.; Chang, S.; Liu, S.; Amini, L.; and Wang, Z. 2020.
\newblock Training stronger baselines for learning to optimize.
\newblock In \emph{Advances in Neural Information Processing Systems}.

\bibitem[{Chen et~al.(2022)Chen, Chen, Cheng, Chen, Awadallah, and Wang}]{chen2022scalable}
Chen, X.; Chen, T.; Cheng, Y.; Chen, W.; Awadallah, A.; and Wang, Z. 2022.
\newblock Scalable learning to optimize: A learned optimizer can train big models.
\newblock In \emph{Proceedings of the European Conference on Computer Vision}, 389--405. Springer.

\bibitem[{Chen et~al.(2024{\natexlab{b}})Chen, Niu, Wang, Xu, Song, and Tan}]{cema}
Chen, Y.; Niu, S.; Wang, Y.; Xu, S.; Song, H.; and Tan, M. 2024{\natexlab{b}}.
\newblock Towards robust and efficient cloud-edge elastic model adaptation via selective entropy distillation.
\newblock In \emph{Proceedings of the International Conference on Learning Representations}.

\bibitem[{Dosovitskiy et~al.(2021)Dosovitskiy, Beyer, Kolesnikov, Weissenborn, Zhai, Unterthiner, Dehghani, Minderer, Heigold, Gelly et~al.}]{vit}
Dosovitskiy, A.; Beyer, L.; Kolesnikov, A.; Weissenborn, D.; Zhai, X.; Unterthiner, T.; Dehghani, M.; Minderer, M.; Heigold, G.; Gelly, S.; et~al. 2021.
\newblock An image is worth 16x16 words: Transformers for image recognition at scale.
\newblock In \emph{Proceedings of the International Conference on Learning Representations}.

\bibitem[{Gandelsman et~al.(2022)Gandelsman, Sun, Chen, and Efros}]{gandelsman}
Gandelsman, Y.; Sun, Y.; Chen, X.; and Efros, A. 2022.
\newblock Test-time training with masked autoencoders.
\newblock In \emph{Advances in Neural Information Processing Systems}, 29374--29385.

\bibitem[{Goyal et~al.(2022)Goyal, Sun, Raghunathan, and Kolter}]{goyal2022test}
Goyal, S.; Sun, M.; Raghunathan, A.; and Kolter, J.~Z. 2022.
\newblock Test time adaptation via conjugate pseudo-labels.
\newblock In \emph{Advances in Neural Information Processing Systems}, volume~35, 6204--6218.

\bibitem[{Hendrycks et~al.(2021{\natexlab{a}})Hendrycks, Basart, Mu, Kadavath, Wang, Dorundo, Desai, Zhu, Parajuli, Guo et~al.}]{ImageNet-R}
Hendrycks, D.; Basart, S.; Mu, N.; Kadavath, S.; Wang, F.; Dorundo, E.; Desai, R.; Zhu, T.; Parajuli, S.; Guo, M.; et~al. 2021{\natexlab{a}}.
\newblock The many faces of robustness: A critical analysis of out-of-distribution generalization.
\newblock In \emph{Proceedings of the IEEE/CVF International Conference on Computer Vision}, 8340--8349.

\bibitem[{Hendrycks and Dietterich(2019)}]{hendrycks2019benchmarking}
Hendrycks, D.; and Dietterich, T. 2019.
\newblock Benchmarking neural network robustness to common corruptions and perturbations.
\newblock In \emph{Proceedings of the International Conference on Learning Representations}.

\bibitem[{Hendrycks et~al.(2021{\natexlab{b}})Hendrycks, Zhao, Basart, Steinhardt, and Song}]{ImageNet-A}
Hendrycks, D.; Zhao, K.; Basart, S.; Steinhardt, J.; and Song, D. 2021{\natexlab{b}}.
\newblock Natural adversarial examples.
\newblock In \emph{Proceedings of the IEEE/CVF Conference on Computer Vision and Pattern Recognition}, 15262--15271.

\bibitem[{Hochreiter and Schmidhuber(1997)}]{lstm}
Hochreiter, S.; and Schmidhuber, J. 1997.
\newblock Long short-term memory.
\newblock \emph{Neural Computation}, 9(8): 1735--1780.

\bibitem[{Iwasawa and Matsuo(2021)}]{T3A}
Iwasawa, Y.; and Matsuo, Y. 2021.
\newblock Test-time classifier adjustment module for model-agnostic domain generalization.
\newblock In \emph{Advances in Neural Information Processing Systems}, volume~34, 2427--2440.

\bibitem[{Khurana et~al.(2021)Khurana, Paul, Rai, Biswas, and Aggarwal}]{khurana2021sita}
Khurana, A.; Paul, S.; Rai, P.; Biswas, S.; and Aggarwal, G. 2021.
\newblock Sita: Single image test-time adaptation.
\newblock \emph{arXiv preprint arXiv:2112.02355}.

\bibitem[{Kingma and Ba(2015)}]{adam}
Kingma, D.~P.; and Ba, J. 2015.
\newblock Adam: {A} method for stochastic optimization.
\newblock In \emph{Proceedings of the International Conference on Learning Representations}.

\bibitem[{Lee et~al.(2024)Lee, Jung, Lee, Park, Shin, Hwang, and Yoon}]{deyo}
Lee, J.; Jung, D.; Lee, S.; Park, J.; Shin, J.; Hwang, U.; and Yoon, S. 2024.
\newblock Entropy is not enough for test-time adaptation: From the perspective of disentangled factors.
\newblock In \emph{Proceedings of the International Conference on Learning Representations}.

\bibitem[{Li et~al.(2020)Li, Chen, You, Wang, and Lin}]{li2020halo}
Li, C.; Chen, T.; You, H.; Wang, Z.; and Lin, Y. 2020.
\newblock Halo: Hardware-aware learning to optimize.
\newblock In \emph{Proceedings of the European Conference on Computer Vision}, 500--518.

\bibitem[{Li and Malik(2016)}]{li2016learning}
Li, K.; and Malik, J. 2016.
\newblock Learning to optimize.
\newblock \emph{arXiv preprint arXiv:1606.01885}.

\bibitem[{Li and Malik(2017)}]{li2017learning}
Li, K.; and Malik, J. 2017.
\newblock Learning to optimize neural nets.
\newblock \emph{arXiv preprint arXiv:1703.00441}.

\bibitem[{Liu et~al.(2021)Liu, Kothari, Van~Delft, Bellot-Gurlet, Mordan, and Alahi}]{TTT++}
Liu, Y.; Kothari, P.; Van~Delft, B.; Bellot-Gurlet, B.; Mordan, T.; and Alahi, A. 2021.
\newblock TTT++: When does self-supervised test-time training fail or thrive?
\newblock In \emph{Advances in Neural Information Processing Systems}, 21808--21820.

\bibitem[{Lv, Jiang, and Li(2017)}]{lv2017learning}
Lv, K.; Jiang, S.; and Li, J. 2017.
\newblock Learning gradient descent: Better generalization and longer horizons.
\newblock In \emph{Proceedings of the International Conference on Machine Learning}, 2247--2255.

\bibitem[{Metz et~al.(2019)Metz, Maheswaranathan, Nixon, Freeman, and Sohl-Dickstein}]{metz2019understanding}
Metz, L.; Maheswaranathan, N.; Nixon, J.; Freeman, D.; and Sohl-Dickstein, J. 2019.
\newblock Understanding and correcting pathologies in the training of learned optimizers.
\newblock In \emph{Proceedings of the International Conference on Machine Learning}, 4556--4565.

\bibitem[{Mirza et~al.(2022)Mirza, Micorek, Possegger, and Bischof}]{DUA}
Mirza, M.~J.; Micorek, J.; Possegger, H.; and Bischof, H. 2022.
\newblock The norm must go on: dynamic unsupervised domain adaptation by normalization.
\newblock In \emph{Proceedings of the IEEE Conference on Computer Vision and Pattern Recognition}, 14765--14775.

\bibitem[{Nado et~al.(2020)Nado, Padhy, Sculley, D'Amour, Lakshminarayanan, and Snoek}]{nado2020evaluating}
Nado, Z.; Padhy, S.; Sculley, D.; D'Amour, A.; Lakshminarayanan, B.; and Snoek, J. 2020.
\newblock Evaluating prediction-time batch normalization for robustness under covariate shift.
\newblock \emph{arXiv preprint arXiv:2006.10963}.

\bibitem[{Niu et~al.(2024)Niu, Miao, Chen, Wu, and Zhao}]{foa}
Niu, S.; Miao, C.; Chen, G.; Wu, P.; and Zhao, P. 2024.
\newblock Test-time model adaptation with only forward passes.
\newblock In \emph{Proceedings of the International Conference on Machine Learning}.

\bibitem[{Niu et~al.(2022)Niu, Wu, Zhang, Chen, Zheng, Zhao, and Tan}]{EATA}
Niu, S.; Wu, J.; Zhang, Y.; Chen, Y.; Zheng, S.; Zhao, P.; and Tan, M. 2022.
\newblock Efficient test-time model adaptation without forgetting.
\newblock In \emph{Proceedings of the International Conference on Machine Learning}, 16888--16905.

\bibitem[{Niu et~al.(2023)Niu, Wu, Zhang, Wen, Chen, Zhao, and Tan}]{SAR}
Niu, S.; Wu, J.; Zhang, Y.; Wen, Z.; Chen, Y.; Zhao, P.; and Tan, M. 2023.
\newblock Towards stable test-time adaptation in dynamic wild world.
\newblock In \emph{Proceedings of the International Conference on Learning Representations}.

\bibitem[{Sun et~al.(2024)Sun, Li, Dalal, Xu, Vikram, Zhang, Dubois, Chen, Wang, Koyejo et~al.}]{sun2024learning}
Sun, Y.; Li, X.; Dalal, K.; Xu, J.; Vikram, A.; Zhang, G.; Dubois, Y.; Chen, X.; Wang, X.; Koyejo, S.; et~al. 2024.
\newblock Learning to (learn at test time): Rnns with expressive hidden states.
\newblock \emph{arXiv preprint arXiv:2407.04620}.

\bibitem[{Sun et~al.(2020)Sun, Wang, Liu, Miller, Efros, and Hardt}]{TTT}
Sun, Y.; Wang, X.; Liu, Z.; Miller, J.; Efros, A.; and Hardt, M. 2020.
\newblock Test-time training with self-supervision for generalization under distribution shifts.
\newblock In \emph{Proceedings of the International Conference on Machine Learning}, 9229--9248.

\bibitem[{Tan et~al.(2024)Tan, Chen, Wu, Zhang, Chen, Zhao, and Niu}]{eata-c}
Tan, M.; Chen, G.; Wu, J.; Zhang, Y.; Chen, Y.; Zhao, P.; and Niu, S. 2024.
\newblock Uncertainty-calibrated test-time model adaptation without forgetting.
\newblock \emph{arXiv preprint arXiv:2403.11491}.

\bibitem[{Vicol, Metz, and Sohl-Dickstein(2021)}]{vicol2021unbiased}
Vicol, P.; Metz, L.; and Sohl-Dickstein, J. 2021.
\newblock Unbiased gradient estimation in unrolled computation graphs with persistent evolution strategies.
\newblock In \emph{Proceedings of the International Conference on Machine Learning}, 10553--10563. PMLR.

\bibitem[{Wang et~al.(2021)Wang, Shelhamer, Liu, Olshausen, and Darrell}]{TENT}
Wang, D.; Shelhamer, E.; Liu, S.; Olshausen, B.; and Darrell, T. 2021.
\newblock Tent: Fully test-time adaptation by entropy minimization.
\newblock In \emph{Proceedings of the International Conference on Learning Representations}.

\bibitem[{Wang et~al.(2019)Wang, Ge, Lipton, and Xing}]{ImageNet-sketch}
Wang, H.; Ge, S.; Lipton, Z.; and Xing, E.~P. 2019.
\newblock Learning robust global representations by penalizing local predictive power.
\newblock In \emph{Advances in Neural Information Processing Systems}, 10506--10518.

\bibitem[{Wang et~al.(2022)Wang, Fink, Van~Gool, and Dai}]{cotta}
Wang, Q.; Fink, O.; Van~Gool, L.; and Dai, D. 2022.
\newblock Continual test-time domain adaptation.
\newblock In \emph{Proceedings of the IEEE/CVF Conference on Computer Vision and Pattern Recognition}, 7201--7211.

\bibitem[{Wen et~al.(2023)Wen, Niu, Li, Wu, Tan, and Wu}]{wen2023test}
Wen, Z.; Niu, S.; Li, G.; Wu, Q.; Tan, M.; and Wu, Q. 2023.
\newblock Test-time model adaptation for visual question answering with debiased self-supervisions.
\newblock \emph{IEEE Transactions on Multimedia}.

\bibitem[{Wichrowska et~al.(2017)Wichrowska, Maheswaranathan, Hoffman, Colmenarejo, Denil, Freitas, and Sohl-Dickstein}]{wichrowska2017learned}
Wichrowska, O.; Maheswaranathan, N.; Hoffman, M.~W.; Colmenarejo, S.~G.; Denil, M.; Freitas, N.; and Sohl-Dickstein, J. 2017.
\newblock Learned optimizers that scale and generalize.
\newblock In \emph{Proceedings of the International Conference on Machine Learning}, 3751--3760.

\bibitem[{Yuan, Xie, and Li(2023)}]{yuan2023robust}
Yuan, L.; Xie, B.; and Li, S. 2023.
\newblock Robust test-time adaptation in dynamic scenarios.
\newblock In \emph{Proceedings of the IEEE/CVF Conference on Computer Vision and Pattern Recognition}, 15922--15932.

\bibitem[{Zeng et~al.(2023)Zeng, Deng, Xu, Niu, and Chen}]{AME}
Zeng, R.; Deng, Q.; Xu, H.; Niu, S.; and Chen, J. 2023.
\newblock Exploring motion cues for video test-time adaptation.
\newblock In \emph{Proceedings of the International Conference on Multimedia}, 1840--1850.

\bibitem[{Zhang, Levine, and Finn(2022)}]{MEMO}
Zhang, M.; Levine, S.; and Finn, C. 2022.
\newblock Memo: Test time robustness via adaptation and augmentation.
\newblock In \emph{Advances in Neural Information Processing Systems}, 38629--38642.

\end{thebibliography}

\section*{Appendices}
In this appendix, we provide more details and more experimental results of our method. We organize the appendix into the following sections.
\begin{itemize}
    \item Section A further introduces the datasets in our experiments.
    \item Section B show more implementation details of our method and compared methods.
    \item Section C provides more experimental results and analyses of our method.
\end{itemize}

\subsection*{A. More Details about Datasets}
\label{sec:dataset}
In our paper, we utilize four variant datasets of ImageNet to validate the effectiveness and generalizability of our proposed method against different types of distribution shifts, including ImageNet-C~\cite{hendrycks2019benchmarking}, ImageNet-R~\cite{ImageNet-R}, ImageNet-Sketch~\cite{ImageNet-sketch}, and ImageNet-A~\cite{ImageNet-A}.

\textbf{ImageNet-C} is derived from the original ImageNet dataset by introducing various types of corruptions. Each type of corruption consists of 50,000 samples spanning 1,000 categories and includes five severity levels.The validation set contains four types of corruptions, namely: Speckle Noise, Spatter, Gaussian Blur, and Saturate. The test set comprises 15 different types of corruptions, namely: Gaussian Noise, Shot Noise, Impulse Noise, Defocus Blur, Glass Blur, Motion Blur, Zoom Blur, Snow, Frost, Fog, Brightness, Contrast, Elastic Transform, Pixelate, and JPEG Compression. We utilized the data from the validation set to train the Meta Gradient Generator (MGG) and conducted TTA using the data from the test set. All corruptions in the dataset are applied at severity level 5.

\textbf{ImageNet-R} encompasses 200 classes from the original ImageNet, consisting of 30,000 images that have been rendered in various artistic styles. These images appear in diverse forms such as sketches, cartoons, sculptures, and other artistic renditions.

\textbf{ImageNet-Sketch} includes 50,000 images corresponding to 1,000 classes from the original ImageNet dataset, with each class containing 50 samples. All images in this dataset are black-and-white sketches.

\textbf{ImageNet-A} contains approximately 7,500 images across 200 classes from the ImageNet dataset. ImageNet-A specifically includes images that are prone to causing errors in model predictions, such as objects captured at unusual angles or with significant occlusion. These challenging characteristics make the dataset particularly adversarial.

\subsection*{B. More Implementation Details}
\label{sec:implementation_detail}
\subsubsection*{B.1. Preprocessing of MGG Input Gradients}
Following previous learning-to-optimize approaches~\cite{lv2017learning, chen2022scalable}, we preprocess the gradient of the model parameters $\theta$ before inputting it into the Meta Gradient Generator (MGG). This preprocessing offers two key advantages. First, by excluding the absolute size of gradients from the optimizer's input, it enhances the optimizer's robustness. Second, this preprocessing can be viewed as a form of normalization. By limiting the input range, it makes the training process for neural network-based MGG models somewhat easier. The preprocessing procedure is similar to that used in Adam~\cite{adam}. Assuming $g_t$ represents the gradient of model parameter $\theta$ at time $t$, the gradient preprocessing process can be expressed as:
\begin{eqnarray}
m = \beta_1 m + (1 - \beta_1) g_t, \quad v = \beta_2 v + (1 - \beta_2) g_t^2 ,
\end{eqnarray}
\begin{eqnarray}
\bar{m} = m / (1 - \beta_1^{t+1}), \quad \bar{v} = v / (1 - \beta_2^{t+1}),
\end{eqnarray}
\begin{eqnarray}
\tilde{g} = g_t / \sqrt{\bar{v} + \epsilon}, \quad \tilde{m} = \bar{m} / \sqrt{\bar{v} + \epsilon},
\end{eqnarray}
\begin{eqnarray}
z_t = \text{concat}(\tilde{g}_t, \tilde{m}_t) ,
\end{eqnarray}
where $\beta_1$ and $\beta_2$ are exponential moving average factors, set to 0.9 and 0.99, respectively. $\tilde{g}$ and $\tilde{m}$ are the scaled gradient and its momentum. After preprocessing $g_t$ to obtain $z_t$, use $z_t$ as the input for MGG to continue the gradient optimization process.

\subsubsection*{B.2. Details about Hyper-parameter Settings}
We use a pre-trained ViT-base~\cite{vit} as our model. This model has been pre-trained on ImageNet-1k, and we load the pre-trained weights using the timm library. We will describe the implementation details of each method.

\textbf{MGTTA (Ours)}. We use Eqn. (7) as the objective function to pre-train the Meta Gradient Generator (MGG) and for TTA, following FOA~\cite{foa}, we set the hyper-parameter $\lambda$ to 0.4. The GML hidden size is set to 8. \textbf{For pre-training MGG}, we randomly select 128 unlabeled samples from the ImageNet-C validation set as its training set. The learning rate is set to 1e-4 for $\theta$ and 1e-2 for $\phi^r$. We update $\theta$ and $\phi^r$ for $T$=2,000 iterations with a batch size of 2. Every 64 iterations, we randomly reinitialize the memory parameters $\phi^m$ of MGG and perform an evaluation on the validation set, ultimately selecting the MGG with the best evaluation results for TTA.
\textbf{During TTA}, we load MGG's pre-trained parameters for $\phi^r$ and keep them fixed, while $\phi^m$ is reinitialized randomly and continuously updated. The learning rate for $\theta$ is set to 1e-3 with a batch size of 64, and 1e-4 with a batch size of 2 or 4.

\textbf{LAME}~\cite{LAME}. Following the hyper-parameter settings of TENT unless not provided, we set the batch size to 64 to maintain consistency with other comparison methods, and we set the value of $k$ in the kNN affinity matrix to 5.

\textbf{T3A}~\cite{T3A}. Following the hyper-parameter settings of T3A unless not provided, we set the batch size to 64, and the number of supports to restore $M$ is set to 20.

\textbf{TENT}~\cite{TENT}. Following the hyper-parameter settings of TENT unless not provided, we use SGD with a momentum of 0.9 as the optimizer, set the batch size to 64, and the learning rate to 0.001.

\textbf{CoTTA}~\cite{cotta}. Following the hyper-parameter settings of CoTTA unless not provided, we use SGD with a momentum of 0.9 as the optimizer, set the batch size to 64, and the learning rate to 0.05, with an augmentation threshold $p_{th}$ of 0.1. If images are below this threshold, we use 32 augmentations. The restoration probability is set to 0.01, and the EMA factor $\alpha$ for teacher update is set to 0.999.

\textbf{SAR}~\cite{SAR}. Following the hyper-parameter settings of SAR unless not provided, we use SGD with a momentum of 0.9 as the optimizer, set the batch size to 64, and the learning rate to 0.001. The entropy threshold $E_0$ is set to $0.4 \times \ln C$, where $C$ is the number of classes.

\textbf{FOA}~\cite{foa}. Following the hyper-parameter settings of SAR unless not provided, we set the batch size to 64, the population size $K$ to 28, the trade-off hyper-parameter $\lambda$ to 0.4, and the number of added prompts to 3.

\textbf{EATA}~\cite{EATA}. Following EATA, we use SGD with a momentum of 0.9 as the optimizer, set the batch size to 64. We set the learning rate to 0.00025. The $E_0$ for reliable sample identification is set to 0.5 and the $\epsilon$ for redundant sample identification to 0.05.

\textbf{DeYO}~\cite{deyo}. Following DeYO, we use SGD with a momentum of 0.9 as the optimizer, set the batch size to 64. We set the learning rate to 0.0005. $Ent_0$ and $\tau_{Ent}$ are set to $0.4 \times \ln C$ and $0.3 \times \ln C$ respectively, where $C$ is the number of classes.

Notably, for LAME, T3A, TENT, CoTTA, SAR, FOA, we strictly follow the hyper-parameters given in the FOA paper, as we use FOA's code as our codebase. 
For EATA and DeYO, since our backbone's pre-trained weights differ from them and their hyper-parameters include threshold settings, which are typically sensitive, we selected better hyper-parameters for these two methods.

\begin{table}[th!]
\centering
\setlength{\tabcolsep}{4pt}
\resizebox{\columnwidth}{!}{
    \begin{tabular}{lcccccccc}
               & NoAdapt & TENT  & SAR  & FOA  & EATA & DeYO & Ours \\ \midrule
    Memory(MB) & 819     & 5,165  & 5,166 & 832  & 5,175 & 5,351 & 5,193 \\
    Runtime(s)        & 54.3                            & 122.6                     & 242.7                & 1636.7               & 127.4                 & 172.4               & 125.5  \\
    
    Acc.(\%)   & 55.5    & 59.6   & 62.7 & 66.3 & 66.7 & 68.2 & 71.3
    \end{tabular}
}
\caption{Efficieny comparison w.r.t Memory Usage(MB) and Wall-clock Runtime(s), which are measured on ImageNet-C test set on a single RTX4090 GPU by processing 50,000 images with a batch size of 64. We report the average accuracy(\%) across 15 corruptions on ImageNet-C(level 5).}
\label{table:memory}
\end{table}

\subsection*{C. More Experimental Results}
\label{sec:more_result}

\subsubsection*{C.1. Run-Time Memory of MGTTA}
Compared to gradient-based methods like TENT, we introduce a new neural network, namely the Meta Gradient Generator (MGG), to optimize and generate better gradients. We compare the memory usage of different TTA methods on the ImageNet-C test set with a batch size of 64. Results in Table \ref{table:memory} indicate that our method only consumes 28MB more memory than TENT (5,193MB vs. 5,165MB) and 158MB less than the previous SOTA DeYO (5,193MB vs. 5,351MB). This demonstrates that introducing MGG only slightly increases memory consumption, yet our method achieves the best accuracy at 71.3\%, significantly outperforming other methods. Compared to FOA, which has the least memory consumption, although our memory usage is greater by 6.7x, experiments about runtime in section 4.4 of the main paper reveal that our method, when updating with all samples, is 13x faster than FOA and achieves superior results (71.3\% vs. 66.3\%).

\subsubsection*{C.2.Effect of GML Network Size}
We use the Gradient Memory Layer(GML) to memorize historical information. Here, we investigate the impact of the GML network size, measured by the number of neurons or hidden size. As shown in Table \ref{table:GML_size}, we conduct experiments under different hidden size. The results indicate that a hidden size of just 2 results in poor memory capacity, but increasing the hidden size enhances this capability. At a hidden size of 8, the accuracy on ImageNet-C reaches an optimal 71.3\%, and on ImageNet-R, it achieves 70.2\%, close to the optimal value. Further increasing the hidden size does not improve performance and additionally increases the storage consumption for memorizing historical information. Ultimately, we selected a hidden size of 8 for GML.
\begin{table}[th!]
\centering
\resizebox{0.9\columnwidth}{!}{
\begin{tabular}{lcccccc}
    Dataset                        & \textit{d}=2    & \textit{d}=4             & \textit{d}=8             & \textit{d}=16   & \textit{d}=32   & \textit{d}=64   \\ \midrule
    ImageNet-C               & 4.1  & 71.2          & \textbf{71.3} & 71.3 & 71.3 & 71.3 \\
    ImageNet-R               & 13.5 & \textbf{70.3} & 70.2          & 70.1 & 70.1 & 70.1
    \end{tabular}
}
\caption{Effect of different \textbf{hidden size (\textit{d})} in gradient memory layer(GML). We report average accuracy (\%) over 15 corruptions of ImageNet-C (level 5).}
\label{table:GML_size}
\end{table}

\begin{table}[th!]
\centering
\setlength{\tabcolsep}{3pt}
\resizebox{\columnwidth}{!}
{
    \begin{tabular}{lccccccccc}
    \multirow{2}{*}{Method} & \multicolumn{9}{c}{Learning Rate}                                                             \\ \cmidrule{2-10} 
                            & 5e-1 & 1e-1 & 5e-2         & 1e-2 & 5e-3 & 1e-3         & 5e-4         & 1e-4 & 5e-5 \\ \midrule
    DeYO                    & 0.2   & 0.4   & 1.0           & 28.9  & 44.9  & 67.9          & \textbf{68.2} & 64.4  & 62.4  \\
    Ours w/o MGG                & 0.2   & 69.2  & \textbf{70.0} & 68.9  & 67.7  & 63.4          & 61.4          & 58.1  & 57.4  \\ 
    Ours                    & 0.2   & 0.4   & 8.9           & 64.1  & 69.4  & \textbf{71.3} & 71.0          & 68.4  & 66.2 
    \end{tabular}

}
\caption{Effect of different learning rates. We report the average accuracy(\%) across 15 corruptions on ImageNet-C(level 5).}
\label{table:learning_rate}
\end{table}

\subsubsection*{C.3. Effect of Learning Rate}
To choose the best leraning rate for our method during TTA and to verify whether the superior performance of our method is due to the optimization ability of MGG on gradients rather than different learning rates, we conducted experiments on ImageNet-C. The results in Table \ref{table:learning_rate} demonstrate that with optimal learning rates, our method outperforms Ours w/o MGG by 1.3\% (71.3\% vs. 70.0\%). This indicates that the performance enhancement is not due to differences in learning rates, but rather because our method uses MGG to optimize and generate better gradients, demonstrating the effectiveness of MGG.

\end{document}